# Bridging the Arithmetic Gap: The Cognitive Complexity Benchmark and Financial-PoT for Robust Financial Reasoning

Boxiang Zhao*, Qince Li, Zhonghao Wang, Yi Wang,
Peng Cheng, and Bo Lin

Tele-Communication Technology Bureau, Xinhua News Agency
zhaobx9676@gmail.com

**Abstract.** While Large Language Models excel at semantic tasks, they face a critical bottleneck in financial quantitative reasoning, frequently suffering from "Arithmetic Hallucinations" and a systemic failure mode we term "Cognitive Collapse". To strictly quantify this phenomenon, we introduce the Cognitive Complexity Benchmark (CCB), a robust evaluation framework grounded in a dataset constructed from 95 real-world Chinese A-share annual reports. Unlike traditional datasets, the CCB stratifies financial queries into a three-dimensional taxonomy, Data Source, Mapping Dificulty, and Result Unit, enabling the precise diagnosis of reasoning degradation in high-cognitive-load scenarios. To address these failures, we propose the Iterative Dual-Phase Financial-PoT framework. This neuro-symbolic architecture enforces a strict architectural decoupling: it first isolates semantic variable extraction and logic formulation, then offloads computation to an iterative, self-correcting Python sandbox to ensure deterministic execution. Evaluation on the CCB demonstrates that while standard Chain-of-Thought falters on complex tasks, our approach offers superior robustness, elevating the Qwen3-235B model's average accuracy from 59.7% to 67.3% and achieving gains of up to 10-fold in high-complexity reasoning tasks. These findings suggest that architectural decoupling is a critical enabling factor for improving reliability in financial reasoning tasks, providing a transferable architectural insight for precision-critical domains that require tight alignment between semantic understanding and quantitative computation.

**Keywords:** Cognitive Complexity Benchmark · Financial-PoT framework · Financial Quantitative Reasoning.

## 1 Introduction

The financial industry relies heavily on the synthesis of information from massive volumes of unstructured data [7]. Annual reports, prospectuses, and earnings calls contain the critical indicators of corporate health. Traditionally, extracting quantitative insights from these documents, such as calculating Free Cash Flow or assessing year-over-year revenue growth, has been a labor-intensive process reserved for skilled analysts. The sheer density of information and the non-standardized formats of OCR-scanned PDF documents create a significant barrier to automation. Consequently, the emergence of Large Language Models (LLMs) has precipitated a transformative shift in FinTech [11], offering the potential to automate these workflows with unprecedented speed and scalability.

Recent advancements have seen the deployment of domain-specific models, such as FinMem [21] and fine-tuned general-purpose LLMs, which have achieved human-level performance on semantic tasks like sentiment analysis and risk classification [12, 16]. However, a glaring deficiency remains: Quantitative Precision. Unlike creative writing, financial analysis is deterministic; the "Net Margin" must be exact. Current paradigms, primarily relying on Chain-of-Thought (CoT) prompting [19], attempt to force LLMs to perform arithmetic operations internally. While effective for logical decomposition, this approach encounters inherent reliability bottlenecks in numerical processing, as LLMs function as probabilistic token predictors rather than symbolic calculators. Consequently, this architectural mismatch manifests as frequent "Arithmetic Hallucinations", where models might confidently assert that 9.11 > 9.9, or fail to maintain temporal consistency across fiscal years.

Beyond simple arithmetic errors, our pilot experiments reveal a more systemic failure mode which we term "Cognitive Collapse". We observe that model performance does not degrade linearly with task dificulty; rather, it collapses precipitously once a complexity threshold is crossed, as shown in Figure 1. To rigorously investigate this phenomenon, we stratified financial tasks into a three-dimensional taxonomy: Data Source, Mapping Dificulty, and Result Unit. Benchmarking three inference paradigms, Direct Generation, CoT, and Program-of-Thought (PoT) [1], within this framework revealed a critical divergence. While Direct and CoT approaches suffer from significant degradation in high-complexity scenarios (such as Cross-table Synthesis), PoT demonstrates remarkable robustness. This finding suggests that the key to preventing "Cognitive Collapse" lies in effectively ofloading computation.

This empirical observation aligns with the cognitive workflows of human financial analysts. When expert analysts answer complex questions, they do not perform long division in their heads. Instead, they strictly separate the semantic task of extracting and transcribing key data points from the symbolic task of executing formulas via external tools. We posit that the limitations of current LLM methodologies, and the root cause of the observed "Cognitive Collapse", arise precisely from the attempt to merge these disparate cognitive functions into a monolithic generation pass.

To bridge this gap and formalize this decoupling, we introduce the Iterative Dual-Phase Financial-PoT framework. Unlike standard approaches, our architecture enforces a strict separation between semantic understanding and symbolic calculation. In the initial Semantic Parsing phase, the LLM functions exclusively as a reader, isolating variables into a structured schema to filter out textual noise. Subsequently, this output is processed in an iterative execution loop, where the schema is translated into Python code to ensure logical consis-tency. By decoupling these tasks, we ensure that the model's reasoning capacity is dedicated solely to document comprehension, leaving the mathematical rigor to the interpreter.

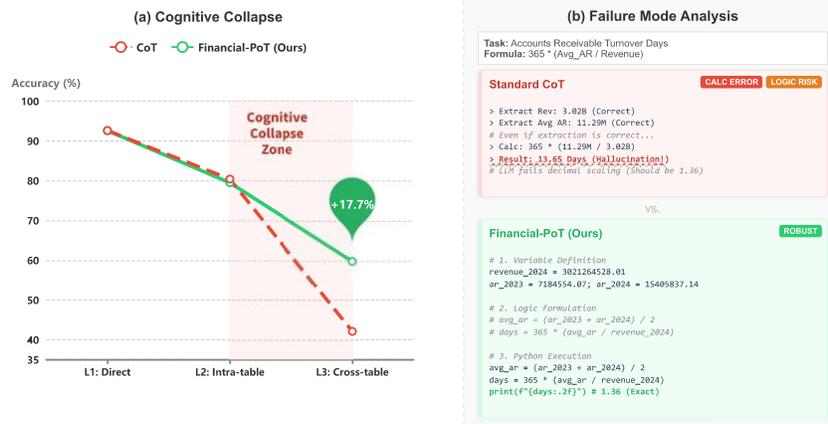

Fig. 1: The "Cognitive Collapse" Phenomenon and the Neuro-Symbolic Solution. (a) Macro Trend: Performance of GPT-oss-120B on the CCB benchmark. While standard CoT (red) degrades non-linearly as complexity increases, plummeting to 42.1% in cross-table synthesis tasks, our Financial-PoT (green) maintains robustness (+17.7% gain). (b) Micro Mechanism: A toy example on Accounts Receivable Turnover Days. Standard CoT suffers from a "Double Failure Mode": it risks retrieving the wrong temporal variable (Logic Risk) and fails decimal scaling during calculation (Arithmetic Hallucination). Financial-PoT resolves this by decoupling semantic variable extraction from symbolic execution within a Python sandbox.

Although our approach is related to PoT and Program-Aided Language models (PAL), it differs in both architectural design and empirical focus. Specifically, we enforce a strict dual-phase decoupling between semantic variable extraction and symbolic computation, combined with an iterative execution-and-correction loop to handle noisy and ambiguous financial documents. Moreover, instead of assuming tool use is universally beneficial, we introduce a dedicated benchmark to systematically analyze when neural-only reasoning collapses and why tool-based execution becomes necessary.

The contributions of this paper are threefold:

- The Cognitive Complexity Benchmark (CCB): A novel evaluation framework is established using real-world, noisy financial documents to simulate industrial complexity. By stratifying cognitive load via a multi-dimensional taxonomy, it explicitly exposes specific logical bottlenecks, serving as a precise diagnostic tool for quantifying reasoning degradation.
- The Financial-PoT Framework and Comparative Analysis: The Financial-PoT framework is proposed, and its eficacy is systematically investigated against Direct Generation and CoT. Findings reveal that while natural lan-guage reasoning falters in quantitative tasks, our neuro-symbolic paradigm exhibits superior robustness and accuracy across diverse complexity levels.
– **Empirical Evidence on Scaling Laws:** Extensive experiments were conducted on the CCB across models ranging from 8B to large-scale proprietary parameters. The results demonstrate that optimizing the reasoning paradigm yields greater performance gains on complex tasks than model scaling alone, offering a cost-effective path for reliable financial AI.

## 2   Related Work

The landscape of financial AI research is rapidly evolving, focusing on the transition to domain-adapted generative paradigms, the integration of neuro-symbolic architectures to mitigate reasoning fragility, and the establishment of multi-dimensional benchmarks for complex cognitive evaluation.

**Language Models in Finance (FinLLMs).** Generative Financial LLMs, such as FinMem [21] and open-source models [8], demonstrate capabilities in semantic tasks but are prone to "arithmetic hallucinations" in quantitative contexts [4]. While recent approaches like Fin-o1 [13] utilize reinforcement learning to mitigate this, they retain a probabilistic nature. In contrast, our work shifts the architectural focus to a deterministic, neuro-symbolic framework to ensure arithmetic precision.

**Chain-of-Thought and Reasoning Limitations.** While Chain-of-Thought (CoT) prompting [19] enhances general mathematical reasoning [15], its reliance on probabilistic token prediction often leads to unfaithful rationales ("semantic mimicking") [10, 17]. In binary-precision financial contexts, this limitation causes error propagation and performance degradation in long-context scenarios [20]. Our study confirms this as "Cognitive Collapse", where performance drops non-linearly as task complexity escalates.

**Neuro-Symbolic AI and Program-Aided Generation.** To mitigate arithmetic fragility, "Tool Learning" paradigms [14] such as PoT [1] and PAL [5] offload computation to external interpreters. Aligning with findings that structured decomposition reduces hallucinations [9], our Financial-PoT advances this by utilizing a Dual-Phase Decoupling Strategy. Unlike end-to-end generation, we strictly separate Semantic Variable Extraction from Syntactic Code Execution to handle real-world noise.

**Financial Reasoning Benchmarks.** Existing benchmarks like FinQA [2] and TAT-QA [22] typically utilize short, curated contexts, while FinanceBench [6] focuses primarily on retrieval. Complementing these, our CCB utilizes contexts composed of several major financial statements extracted from annual reports. Akin to the reasoning depth analysis in GSM8K [3], it stratifies tasks by multi-dimensional cognitive load to precisely diagnose reasoning failures.

Table 1: Summary of Target Financial Indicators and Cognitive Complexity Classification[1]

| Indicator | Formula | Source | CCB Classification (Unit / Source / Dificulty) |
|---|---|---|---|
| **Profitability** | | | |
| ROE | $\frac{\text{Net Inc.}}{\text{Avg. Parent Eq.}}$ | IS + BS | Percentage / Cross-table / Ambiguous |
| ROA | $\frac{\text{Net Inc.}}{\text{Avg. Tot. Assets}}$ | IS + BS | Percentage / Cross-table / Implicit |
| Gross Margin | $\frac{\text{Rev.} - \text{COGS}}{\text{Rev.}}$ | IS Only | Percentage / Intra-table / Implicit |
| Net Margin | $\frac{\text{Net Inc.}}{\text{Rev.}}$ | IS Only | Percentage / Intra-table / Implicit |
| **Solvency** | | | |
| Debt Ratio | $\frac{\text{Tot. Liab.}}{\text{Tot. Assets}}$ | BS Only | Percentage / Intra-table / Implicit |
| Current Ratio | $\frac{\text{Cur. Assets}}{\text{Cur. Liab.}}$ | BS Only | Ratio / Intra-table / Implicit |
| Quick Ratio | $\frac{\text{Cur. Assets} - \text{Inv.}}{\text{Cur. Liab.}}$ | BS Only | Ratio / Intra-table / Implicit |
| **Eficiency** | | | |
| Asset Turnover | $\frac{\text{Rev.}}{\text{Avg. Tot. Assets}}$ | IS + BS | Ratio / Cross-table / Implicit |
| Inv. Days | $\frac{365}{\text{COGS}/\text{Avg. Inv.}}$ | IS + BS | Days / Cross-table / Multi-step |
| AR Days | $\frac{365}{\text{Rev.}/\text{Avg. AR}}$ | IS + BS | Days / Cross-table / Multi-step |
| **Growth** | | | |
| Rev. Growth | $\frac{\text{Cur. Rev.} - \text{Prev. Rev.}}{\text{Prev. Rev.}}$ | IS (Temp.) | Percentage / Intra-table / Implicit |
| Net Profit Gr. | $\frac{\text{Cur. Inc.} - \text{Prev. Inc.}}{\text{Prev. Inc.}}$ | IS (Temp.) | Percentage / Intra-table / Implicit |
| **Cash Flow** | | | |
| OCF | (Direct Extraction) | CFS Only | Currency / Direct / Explicit |
| FCF | OCF − CAPEX | CFS Only | Currency / Intra-table / Ambiguous |

## 3  The Cognitive Complexity Benchmark (CCB)

To rigorously diagnose the mechanics of "Cognitive Collapse", we propose a multi-dimensional taxonomy applied to 14 key financial indicators, as detailed in Table 1. These indicators represent the fundamental metrics used in professional analysis to assess corporate solvency, profitability, and operational eficiency. Unlike traditional benchmarks that rely on a single dificulty metric, the CCB stratifies these queries along three distinct dimensions: Data Source, Mapping Dificulty, and Result Unit. This structure allows us to pinpoint exactly where the model's reasoning process fractures.

---

[1] Abbreviations in Table 1. Indicators: ROE = Return on Equity; ROA = Return on Assets; Inv. Days = Inventory Turnover Days; AR Days = Accounts Receivable Turnover Days; OCF = Operating Cash Flow; CAPEX = Capital Expenditures; FCF = Free Cash Flow. Statements: IS = Income Statement; BS = Balance Sheet; CFS = Cash Flow Statement. Variables: Rev. = Revenue; Gr. = Growth; Inc. = Income; Eq. = Equity; Liab. = Liabilities; Inv. = Inventory; AR = Accounts Receivable; COGS = Cost of Goods Sold; Tot. = Total. Logic: Cur. = Current Period; Prev. = Previous Period; Avg. = Average balances; Temp. = Temporal extraction.

### 3.1   Dimension 1: Data Source (Contextual Scope)

This dimension measures the "search distance" required to locate variables, directly determining the model's working memory load and context retention capability.

- Direct Extraction: The value exists explicitly in the text or table (e.g., OCF). No calculation or memory retention is required.
- Intra-table Calculation: All constituent variables reside within a single financial statement (e.g., Current Ratio relies solely on the Balance Sheet). The context is localized, minimizing noise.
- Cross-table Calculation: Variables are scattered across disparate statements (e.g., ROE requires linking the Income Statement and Balance Sheet). This forces the model to perform "context switching" and temporal alignment (e.g., matching "Current Year" profit with "Average" equity), significantly increasing the risk of hallucination.

### 3.2   Dimension 2: Mapping Dificulty (Semantic Gap)

This dimension evaluates the semantic gap between the natural language query and the standardized line items in the report.

- Explicit Keys: The query term matches the report line item verbatim (e.g., "Find Operating Income").
- Implicit & Multi-step Logic: The formula is standard, but the model must either identify constituent variables (e.g., Gross Margin) or perform chained calculations (e.g., Turnover Days requires deriving an average before division).
- Ambiguous Keys: The query involves non-standardized terms or derivative metrics lacking a direct line item. For instance, calculating Free Cash Flow (FCF) often requires semantically mapping "Cash paid for fixed assets" to "CAPEX", a common source of retrieval failure.

### 3.3   Dimension 3: Result Unit (Output Format)

This dimension evaluates the model's ability to handle distinct numerical schemas and precision requirements inherent to different financial indicators.

- Percentage (%): Applied to profitability and growth metrics (e.g., ROE). Represents normalized proportions or rates of change, requiring the model to derive precise fractional values distinct from structural multiples.
- Ratio (Times): Pure decimal values common in solvency analysis (e.g., Current Ratio). Represents relative multiples or coverage capacities, testing the model's adherence to decimal magnitude without unit conversion.
- Days: Time-based metrics (e.g., Turnover Days). Acts as a stress test by enforcing the use of a temporal constant (365) in the numerator.
- Currency (RMB): Absolute values (e.g., Net OCF). Evaluates the capacity to preserve precision with large-magnitude integers.

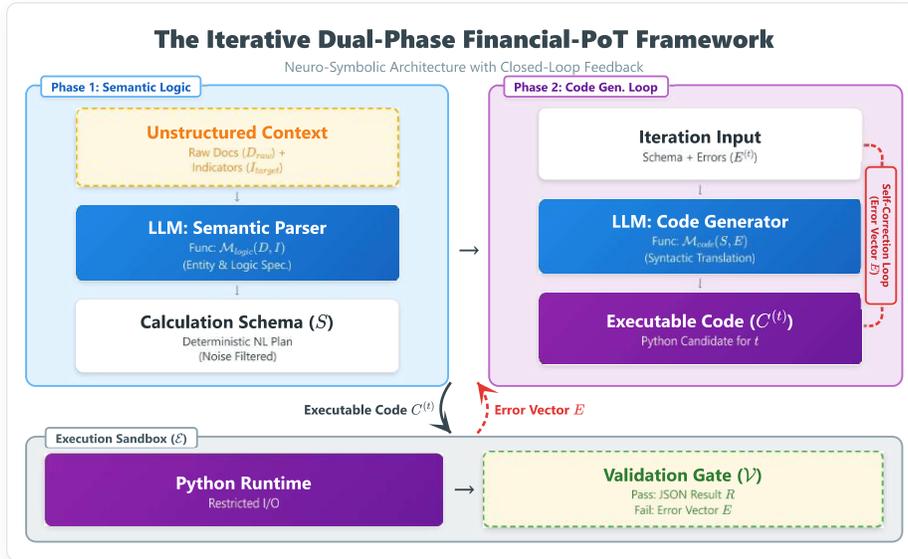

Fig. 2: Financial-PoT Framework.

## 4 The Financial-PoT Framework

We propose the Iterative Dual-Phase Financial-PoT, a neuro-symbolic architecture designed to resolve the dichotomy between semantic ambiguity and quantitative rigor in financial analysis. Unlike standard open-loop generation, our framework enforces an architectural decoupling of Semantic Planning from Syntactic Execution, augmented by a deterministic feedback loop for autonomous self-correction.

### 4.1 Phase 1: Semantic Logic Formulation

The initial phase isolates variable extraction from arithmetic computation. We define a transformation function $M_{logic}$ that maps the unstructured financial context $D_{raw}$ and the set of target indicators $I_{target}$ to a deterministic Calculation Schema ($S_{schema}$).

$$S_{schema} = M_{logic}(D_{raw}, I_{target}) \tag{1}$$

In this semantic parsing stage, the LLM functions exclusively as a domain analyst. It is tasked with two specific objectives: (1) Entity Normalization, where textual representations of magnitude (e.g., "10 billion") are converted into standard numerical literals; and (2) Logic Specification, where the arithmetic relationships between extracted entities are defined in natural language. This intermediate representation serves as a high-fidelity noise filter, reducing the input context to a structured set of variables and formulas $S_{schema}$ prior to code synthesis.

### 4.2 Phase 2: Iterative Code Generation and Self-Correction

The second phase executes the transition from natural language reasoning to symbolic computation. We introduce an iterative refinement loop to mitigate the stochastic fragility inherent in generative models.

Let $C^{(t)}$ denote the executable Python code generated at iteration t. The initial code candidate $C^{(0)}$ is synthesized based on the cleaned schema $S_{schema}$:

$$C^{(0)} = M_{code}(S_{schema}) \tag{2}$$

To ensure robustness, the execution environment E operates within a feedback-driven protocol. At each step t, the execution yields either a valid result set R or an exception vector E:

$$R^{(t)}, E^{(t)} = E(C^{(t)}) \tag{3}$$

The validation function $V(R^{(t)})$ evaluates the completeness and validity of the results (e.g., checking for null values or logical inconsistencies). If $V(R^{(t)})$ fails or if $E^{(t)}$ is non-empty (indicating runtime errors such as ZeroDivisionError or SyntaxError), the system triggers a self-correction mechanism. The exception vector $E^{(t)}$ is fed back into the model to generate a rectified code version $C^{(t+1)}$:

$$C^{(t+1)} = M_{code}(C^{(t)}, E^{(t)}, S_{schema}) \tag{4}$$

This cycle repeats until a valid result is obtained or a maximum iteration depth T is reached (where T = 3). This closed-loop mechanism effectively allows the model to "debug" its own logic dynamically, converting runtime failures into learning signals.

## 5   Experiments

To validate the Financial-PoT framework and rigorously investigate "Cognitive Collapse", we benchmark Direct, CoT, and PoT paradigms using the CCB derived from 95 real-world A-share reports.

### 5.1 Dataset and Experimental Setup

We constructed a specialized benchmark derived from the Annual Reports of 19 listed companies in the Chinese A-share Baijiu industry over the past five years. The raw documents were processed using DeepSeek OCR [18] to parse and extract six major financial statements (Consolidated and Parent Company Balance Sheets, Income Statements, and Cash Flow Statements), which served as the model input. The evaluation focuses on the calculation of 14 key financial indicators: ROE, ROA, Gross Margin, Net Margin, Debt Ratio, Current Ratio, Quick Ratio, Asset Turnover, Inventory Turnover Days, Accounts Receivable Turnover Days, Revenue Growth, Net Profit Growth, OCF, and FCF.

Table 2: Results By Source (Calc Path)

| Category | Model<br>Metric | Qwen3-235B | Qwen3-32B | Qwen3-8B | GPT-oss-120B |
|---|---|---|---|---|---|
| Direct Extraction | Direct | 89.5 | 92.6 | 84.2 | 92.6 |
|  | CoT | 92.6 | 92.6 | 88.4 | 92.6 |
|  | PoT | 90.5 | 68.4 | 66.3 | 92.6 |
| Intra-table Calc | Direct | 64.3 | 45.8 | 20.0 | 79.7 |
|  | CoT | 74.6 | 51.1 | 24.5 | 80.4 |
|  | PoT | 79.7 | 60.4 | 37.6 | 78.0 |
| Cross-table Calc | Direct | 6.9 | 6.3 | 1.7 | 37.9 |
|  | CoT | 29.3 | 12.6 | 1.7 | 42.1 |
|  | PoT | 42.7 | 26.5 | 2.1 | 59.8 |

To evaluate performance across different architectures and parameter scales, we conducted experiments on a diverse set of models. Specifically, we tested the Qwen3 series (covering 8B, 32B, and 235B parameters) and GPT-oss-120B, which was included to serve as a high-performance open-source baseline for cross-architecture validation. We also conducted preliminary evaluations on Fin-o1 [20], a recent model optimized via supervised fine-tuning and reinforcement learning. However, as it exhibited performance comparable to, yet slightly lower than, the Qwen3-8B baseline on our benchmark, these results are omitted from the main discussion to focus on competitive baselines.

### 5.2   Performance Analysis by Data Source (Contextual Scope)

Table 2 examines model performance across varying degrees of contextual dispersion, verifying the correlation between "search distance" and reasoning degradation. The results exhibit a stark polarization: while Direct Extraction tasks achieve near-saturation performance (e.g., 92.6% for Qwen3-235B in CoT mode), accuracy plummets in Cross-table Calculation scenarios, confirming the high cognitive cost of context switching.

In the Cross-table category, which requires linking variables from disparate financial statements (e.g., Income Statement and Balance Sheet), the limitation of the Direct Generation paradigm is most pronounced. Qwen3-235B, despite its scale, scores only 6.9% in Direct mode, indicating a systemic failure in maintaining variable consistency across long contexts. However, the Financial-PoT framework demonstrates exceptional utility here. By explicitly separating variable retrieval from calculation, PoT boosts Qwen3-235B's performance to 42.7%, and enables GPT-oss-120B to reach 59.8%. This suggests that the "Cognitive Collapse" observed in semantic models is largely driven by the inability to align temporal and structural contexts simultaneously, a bottleneck that symbolic execution effectively bypasses.

Conversely, for Direct Extraction and simple Intra-table tasks, the performance gap between paradigms narrows. Notably, in purely extractive tasks, standard CoT occasionally outperforms PoT, a phenomenon predominantly observed in smaller models (e.g., Qwen3-32B drops from 92.6% in CoT to 68.4% in PoT). This indicates that for low-complexity retrieval where no arithmetic logic is required, the overhead of code generation may introduce unnecessary noise that disproportionately impacts smaller architectures. In contrast, GPT-oss-120B maintains comparable results across Direct, CoT, and PoT paradigms in these low-complexity domains, implying superior semantic robustness in basic information retrieval.

Table 3: Results By Dificulty (Key Mapping)

| Category | Model Metric | Qwen3-235B | Qwen3-32B | Qwen3-8B | GPT-oss-120B |
|---|---|---|---|---|---|
| Explicit Keys | Direct | 89.5 | 92.6 | 84.2 | 92.6 |
| | CoT | 92.6 | 92.6 | 88.4 | 92.6 |
| | PoT | 90.5 | 68.4 | 66.3 | 92.6 |
| Implicit Calc | Direct | 50.0 | 38.3 | 16.6 | 64.2 |
| | CoT | 60.9 | 42.9 | 18.5 | 66.6 |
| | PoT | 66.1 | 48.4 | 26.9 | 73.7 |
| Ambiguous Keys | Direct | 16.5 | 4.9 | 0.7 | 61.8 |
| | CoT | 44.6 | 14.0 | 6.3 | 62.5 |
| | PoT | 63.5 | 43.9 | 14.0 | 62.1 |

### 5.3 Performance Analysis by Mapping Dificulty (Semantic Gap)

Table 3 elucidates the impact of the semantic gap between natural language queries and standardized financial line items. The results underscore that Ambiguous Keys, where the target indicator (e.g., FCF) lacks a direct verbatim match in the source text, constitute a primary source of retrieval failure.

In the Ambiguous Keys category, the Direct Generation paradigm collapses significantly. For instance, Qwen3-235B achieves only 16.5% accuracy, suggesting that without intermediate reasoning, models struggle to implicitly map non-standard terms. Standard CoT provides a substantial recovery (improving Qwen3-235B to 44.6%) by allowing the model to "talk through" the semantic alignment. However, the Financial-PoT framework yields the most robust performance, elevating the accuracy to 63.5%. This 18.9% gain over CoT indicates that defining variables within a programming schema ($S_{schema}$) enforces a stricter semantic discipline than free-text reasoning, effectively grounding ambiguous terms into executable logic.

In contrast, Explicit Keys present a minimal cognitive barrier. Most models achieve near-saturation performance (e.g., >90%) regardless of the prompting strategy. Consistent with previous observations, applying PoT to these trivial mappings can be counterproductive for smaller models (e.g., Qwen3-8B drops from 88.4% in CoT to 66.3% in PoT), as the overhead of code syntax generation outweighs the benefits when no semantic disambiguation is required. Meanwhile,

Table 4: Results By Unit (Format)

| Category | Metric | Qwen3-235B | Qwen3-32B | Qwen3-8B | GPT-oss-120B |
|---|---|---|---|---|---|
| Currency (RMB) | Direct | 59.5 | 49.5 | 42.6 | 88.9 |
| | CoT | 72.6 | 54.7 | 49.5 | 90.0 |
| | PoT | 80.5 | 59.5 | 45.3 | 90.0 |
| Days | Direct | 3.7 | 0.5 | 2.6 | 20.0 |
| | CoT | 18.9 | 5.8 | 1.1 | 21.1 |
| | PoT | 35.3 | 16.3 | 4.2 | 46.8 |
| Percentage (%) | Direct | 57.1 | 42.7 | 20.2 | 70.5 |
| | CoT | 66.9 | 47.1 | 21.2 | 72.3 |
| | PoT | 68.6 | 52.0 | 32.0 | 76.8 |
| Ratio (Times) | Direct | 37.5 | 30.5 | 7.0 | 69.5 |
| | CoT | 61.4 | 37.9 | 14.4 | 72.6 |
| | PoT | 76.8 | 56.1 | 18.2 | 68.1 |

Implicit Calculation tasks sit between these extremes, further validating that model performance is inversely proportional to the semantic distance between the query and the data source.

### 5.4 Performance Analysis by Result Unit (Output Format)

Table 4 presents the evaluation results stratified by the output unit format, revealing how the rigidity of arithmetic constraints impacts model reasoning. Consistent with our "Cognitive Collapse" hypothesis, we observe a distinct performance hierarchy: Currency > Percentage ≈ Ratio ≫ Days. The Currency category, which primarily involves absolute value extraction or simple additive operations (e.g., OCF), exhibits the highest baseline accuracy across all models (e.g., 88.9% for GPT-oss-120B in Direct mode).

However, a severe degradation is observed in the Days category (e.g., Inventory Turnover Days), which represents the peak of computational complexity. These indicators require multi-step logic: aggregating cross-table data to calculate an average balance followed by a division operation. Under the Direct Generation paradigm, models suffer a catastrophic failure; notably, even the Qwen3-235B model achieves only 3.7% accuracy, while smaller models fail almost completely. The introduction of the PoT paradigm significantly mitigates this deficiency. By offloading the complex arithmetic logic to the Python interpreter, Qwen3-235B achieves a nearly 10-fold improvement (rising to 35.3%), and GPT-oss-120B improves from 20.0% to 46.8%. Similarly, for Ratio and Percentage tasks, which demand precise division, PoT consistently outperforms CoT. For instance, in the Ratio category, Qwen3-235B sees a substantial gain from 61.4% (CoT) to 76.8% (PoT). These findings confirm that while LLMs struggle with the deterministic rigor required for time-based and ratio-based metrics, architectural decoupling via Financial-PoT effectively bridges this arithmetic gap.

### 5.5  Overall Results

Table 5 provides a granular breakdown of model performance across 14 specific financial indicators, offering conclusive evidence for the "Cognitive Collapse" hypothesis and the eficacy of the Financial-PoT framework. The data reveals a stark performance dichotomy dictated by the structural complexity of the target variable.

**High-Complexity Synthesis (The "Collapse" Zone):** For indicators requiring cross-table synthesis and temporal alignment, such as ROE and AR Turnover Days, the Direct Generation paradigm proves functionally obsolete. For instance, on the ROE task, which necessitates linking Net Income (Income Statement) with Average Equity (Balance Sheet), Qwen3-235B achieves a negligible 6.3% accuracy in Direct mode. Even Standard CoT struggles (31.6%) due to the high probability of hallucinating values during the "context switching" process. However, Financial-PoT restores capability, elevating Qwen3-235B to 38.9% and GPT-OSS-120B to 58.9%. A similar trend is observed in AR Turnover Days, where PoT boosts Qwen3-235B's performance from 5.3% (Direct) to 67.4%, effectively bridging the arithmetic gap through symbolic execution.

**Ambiguity and Multi-step Logic:** The robustness of PoT is most visible in indicators involving ambiguous mappings or subtractive logic, such as the Quick Ratio (Current Assets - Inventory / Current Liabilities) and FCF. In Direct mode, models frequently fail to deduct Inventory correctly, resulting in a 13.7% score for Qwen3-235B on Quick Ratio. The explicit variable definition enforced by PoT corrects this, resulting in a massive leap to 81.1% accuracy. This confirms that code generation acts as a "syntax check" for financial logic, forcing the model to resolve ambiguities before calculation.

**Low-Complexity Retrieval (The Ceiling Effect):** Conversely, for "retrieval-heavy" metrics like OCF and Debt Ratio, the performance gap vanishes. Direct Generation yields near-perfect results (e.g., 90.5% for Qwen3-235B on Debt Ra-tio), and applying PoT often results in a slight regression (dropping to 87.4%) due to the unnecessary overhead of code synthesis. This suggests that a dynamic routing mechanism, which dispatches simple tasks to Direct/CoT and complex tasks to PoT, would be the optimal deployment strategy.

**Overall Trends:** The Total Average scores conclusively demonstrate the superiority of the neuro-symbolic approach for general financial reasoning. Across the 14 indicators, Qwen3-235B improves from an average of 45.6% (Direct) to 67.3% (PoT), and even the smaller Qwen3-32B sees a gain from 35.0% to 48.9%. Notably, the Qwen3-32B model with PoT (48.9%) outperforms the significantly larger Qwen3-235B using Direct generation (45.6%), providing strong empirical evidence that optimizing the reasoning paradigm is more parameter-eficient than model scaling. However, limitations exist at the extremes. For the 8B model, parameter scaling yields greater gains than PoT, highlighting a dual bottleneck where symbolic execution depends on a minimum threshold of semantic understanding. Conversely, GPT-oss-120B shows smaller relative improvements, reflecting diminishing marginal returns due to its already high intrinsic baseline. While certain hyper-complex metrics like Inventory Turnover Days remain challenging for all models (peaking at 25.3%), the Financial-PoT framework successfully establishes a new baseline for reliability in automated financial analysis.

Table 5: Results for Detailed Indicators

| Category | Model Metric | Qwen3-235B | Qwen3-32B | Qwen3-8B | GPT-oss-120B |
|---|---|---|---|---|---|
| Net Profit Growth | Direct | 70.5 | 29.5 | 13.7 | 62.1 |
|  | CoT | 70.5 | 33.7 | 18.9 | 61.1 |
|  | PoT | 81.1 | 62.1 | 40.0 | 47.4 |
| Return on Equity (ROE) | Direct | 6.3 | 8.4 | 1.1 | 42.1 |
|  | CoT | 31.6 | 13.7 | 1.1 | 44.2 |
|  | PoT | 38.9 | 26.3 | 1.1 | 58.9 |
| Inventory Turnover Days | Direct | 2.1 | 0.0 | 0.0 | 2.1 |
|  | CoT | 2.1 | 3.2 | 1.1 | 4.2 |
|  | PoT | 3.2 | 2.1 | 0.0 | 25.3 |
| Accounts Receivable Turnover Days | Direct | 5.3 | 1.1 | 5.3 | 37.9 |
|  | CoT | 35.8 | 8.4 | 1.1 | 37.9 |
|  | PoT | 67.4 | 30.5 | 8.4 | 68.4 |
| Asset Turnover | Direct | 10.5 | 10.5 | 0.0 | 56.8 |
|  | CoT | 42.1 | 22.1 | 2.1 | 68.4 |
|  | PoT | 61.1 | 44.2 | 0.0 | 72.6 |
| Return on Assets (ROA) | Direct | 10.5 | 11.6 | 2.1 | 50.5 |
|  | CoT | 34.7 | 15.8 | 3.2 | 55.8 |
|  | PoT | 43.2 | 29.5 | 1.1 | 73.7 |
| Current Ratio | Direct | 88.4 | 81.1 | 21.1 | 93.7 |
|  | CoT | 92.6 | 80.0 | 33.7 | 93.7 |
|  | PoT | 88.4 | 69.5 | 37.9 | 91.6 |
| Operating Cash Flow (OCF) | Direct | 89.5 | 92.6 | 84.2 | 92.6 |
|  | CoT | 92.6 | 92.6 | 88.4 | 92.6 |
|  | PoT | 90.5 | 68.4 | 66.3 | 92.6 |
| Free Cash Flow (FCF) | Direct | 29.5 | 6.3 | 1.1 | 85.3 |
|  | CoT | 52.6 | 16.8 | 10.5 | 87.4 |
|  | PoT | 70.5 | 50.5 | 24.2 | 87.4 |
| Revenue Growth | Direct | 88.4 | 66.3 | 26.3 | 91.6 |
|  | CoT | 89.5 | 68.4 | 27.4 | 91.6 |
|  | PoT | 87.4 | 67.4 | 49.5 | 91.6 |
| Debt Ratio | Direct | 90.5 | 86.3 | 70.5 | 93.7 |
|  | CoT | 92.6 | 91.6 | 61.1 | 93.7 |
|  | PoT | 87.4 | 67.4 | 57.9 | 93.7 |
| Quick Ratio | Direct | 13.7 | 0.0 | 0.0 | 57.9 |
|  | CoT | 49.5 | 11.6 | 7.4 | 55.8 |
|  | PoT | 81.1 | 54.7 | 16.8 | 40.0 |
| Net Margin | Direct | 52.6 | 55.8 | 24.2 | 61.1 |
|  | CoT | 60.0 | 60.0 | 25.3 | 68.4 |
|  | PoT | 52.6 | 43.2 | 17.9 | 80.0 |
| Gross Margin | Direct | 81.1 | 41.1 | 3.2 | 92.6 |
|  | CoT | 89.5 | 46.3 | 11.6 | 91.6 |
|  | PoT | 89.5 | 68.4 | 56.8 | 92.6 |
| Total Average | Direct | 45.6 | 35.0 | 18.0 | 65.7 |
|  | CoT | 59.7 | 40.3 | 20.9 | 67.6 |
|  | PoT | 67.3 | 48.9 | 27.0 | 72.6 |

## 6  Conclusion and Future Work

In this study, we aimed to mitigate the "Cognitive Collapse" phenomenon in financial analytics by introducing the Cognitive Complexity Benchmark (CCB) and the Iterative Dual-Phase Financial-PoT framework. Our neuro-symbolic approach, which decouples reasoning from calculation, alleviates arithmetic hallucinations to a significant extent where standard paradigms fail. Extensive experiments demonstrate that optimizing the reasoning paradigm often yields greater gains than model scaling, confirming that reliable AI requires symbiotic neuro-symbolic systems rather than merely larger stochastic predictors. Beyond finance, this architectural decoupling establishes a generalized blueprint for precision-critical domains like scientific modeling and supply chain logistics, where the rigorous alignment of semantic understanding and quantitative exactitude is paramount.

Future work will focus on expanding the CCB from 95 reports to a diverse large-scale corpus to test generalization, and integrating dynamic routing to selectively apply PoT for optimal eficiency.